\RequirePackage{fix-cm}
\documentclass{svjour3}                     
\smartqed  

\usepackage{graphicx}
\usepackage{url}
\usepackage{multirow}
\usepackage{float}
\usepackage{tabularx}
\usepackage{makecell}
\usepackage{tabulary}
\usepackage{chngpage}
\usepackage{fancyhdr}
\usepackage{multirow}
\usepackage{amsfonts}
\usepackage{hyperref}
\usepackage{microtype}
\usepackage{array}
\usepackage{tikz}
 
\newcolumntype{?}[1]{!{\vrule width #1}}

\newcommand{\vdir}[1]{\rotatebox[origin=c]{90}{#1}}
\newcommand{\bd}[1]{\textbf{#1}}

\begin{document}

\title{Weakly Supervised Convolutional LSTM Approach for Tool Tracking in Laparoscopic Videos}
\subtitle{}
\titlerunning{Surgical Tool Tracking without Spatial Annotations} 
\author{Chinedu Innocent Nwoye        \and
        Didier Mutter  \and
        Jacques Marescaux  \and
        Nicolas Padoy
}
\institute{Chinedu Innocent Nwoye \and Nicolas Padoy \at
              ICube, University of Strasbourg, CNRS, IHU Strasbourg, France
              \email{nwoye.chinedu@gmail.com . npadoy@unistra.fr}          
         \and 
         Didier Mutter \and Jacques Marescaux \at
              University Hospital of Strasbourg, IRCAD, IHU Strasbourg, France
}
\date{
} 

\maketitle

\begin{abstract}:\\ 
\emph{Purpose}: Real-time surgical tool tracking is a core component of the future intelligent operating room (OR), because it is highly instrumental to analyze and understand the surgical activities. 
Current methods for surgical tool tracking in videos need to be trained on data in which the spatial positions of the tools are manually annotated. Generating such training data is difficult and time-consuming. Instead, we propose to use solely binary presence annotations to train a tool tracker for laparoscopic videos.\\
\emph{Methods}: The proposed approach is composed of a CNN + Convolutional LSTM (\emph{ConvLSTM}) neural network trained end-to-end, but weakly supervised on tool binary presence labels only. 
We use the ConvLSTM to model the temporal dependencies in the motion of the surgical tools and leverage its spatio-temporal ability to smooth the class peak activations in the localization heat maps (\emph{Lh-maps}).\\ 
\emph{Results:} We build a baseline tracker on top of the CNN model and demonstrate that our approach based on the ConvLSTM outperforms the baseline in tool presence detection, spatial localization, and motion tracking by over $5.0\%$, $13.9\%$, and $12.6\%$, respectively.\\ 
\emph{Conclusions:} In this paper, we demonstrate that binary presence labels are sufficient for training a deep learning tracking model using our proposed method. We also show that the ConvLSTM can leverage the spatio-temporal coherence of consecutive image frames across a surgical video to improve tool presence detection, spatial localization, and motion tracking.
\end{abstract}

\section{Introduction}
\label{sec:introduction}
The automated analysis of surgical workflow can support many routine surgical activities by providing clinical decision support, report generation, and data annotation.
This has sparked active research in the medical computer vision community, particularly on surgical phase recognition \cite{tmi:twinanda2017endonet,ipcai:zisimopoulos2018deepphase} and tool detection \cite{ipcai:richa2011visual,miccai:sznitman2014fast,miccai:vardazaryan2018weakly,miccai:sznitman2011unified,mai:al2018monitoring,wacv:jin2018tool}.
Since surgical activities can now be captured using cameras, large amounts of data become available for their analysis. 
Surgical tools tracking is a multi-object tracking (MOT) problem that entails the modeling of the trajectories of all surgical tools throughout a surgical video sequence.
It is needed to model and analyze tool-tissue interactions.
The predominant MOT approach has been \textit{tracking-by-detection} \cite{ipcai:richa2011visual,iccv:singh2017hide,lstm_da:milan2017online}, which is an integration of a detection model, a localization model and a tracking algorithm. 
In this approach, object detectors like \cite{cvpr:he2016deep} are used for predicting the presence or absence of objects of interest. 
The bounding box coordinates of the detected objects are then extracted using a localization model as seen in \cite{miccai:vardazaryan2018weakly,wacv:jin2018tool}. 
Most times, the localization model is regressed over the bounding box annotations in a fully supervised manner. 
This is usually concluded by a one-to-one assignment of the detected objects to object trajectories using a data association algorithm. 
Bipartite graph matching \cite{hungarian:kuhn1955hungarian} has been widely used in this regard. 
Most works in the medical computer vision community view this matching as a linear assignment problem learnable by stochastic optimization \cite{ipcai:richa2011visual,miccai:sznitman2014fast,miccai:sznitman2011unified,cvpr:mishra2017learning}.
Meanwhile, recent works have also shown that the long short-term memory (LSTM) model has the capability to learn a data association task \cite{lstm_da:milan2017online}, making it easier to build a unified deep learning tracking model.

Surgical tool tracking in endoscopic videos is not an easy task. 
In particular, laparoscopic data presents several challenges, such as the presence of blood stains on the tools, smoke from electric coagulation and cutting, motion blur for fast-moving tools, and the removal and re-insertion of the endoscope during the procedure. 
Furthermore, most endoscopic datasets are not fully exploited by deep learning methods because only a small fraction of the dataset can be spatially annotated with localization information. 
The implication is that most intriguing tasks are only explored and tested on a very tiny fraction of the dataset.
Creating spatial annotations such as region boundaries and pixel-wise masks is indeed tedious and time-consuming.
Since generating binary annotations just indicating the presence of the tools requires less effort, exploiting this information for tracking becomes an interesting research question.

Previous tool tracking work in the medical computer vision community relies on spatially annotated data \cite{ipcai:richa2011visual,miccai:sznitman2011unified}.
In this paper, we propose a new deep learning object tracking method that circumvents the lack of spatially annotated surgical data with \textit{weak supervision} on binary presence labels. 
Weak supervision is here motivated by the idea that when a convolutional neural network (CNN) is trained in a fully convolutional manner for a classification task, some of the convolution layers before the dense layer learn a general notion about the detected object. 
The activations in these inner layers can therefore be exploited for other tasks than the ones they were originally trained for. 
Based on this observation, weak supervision has been employed for cancerous region detection \cite{miccai:hwang2016self,ieee:jia2017constrained}, surgical tool center localization \cite{miccai:vardazaryan2018weakly} and object instance segmentation \cite{arXiv:zhou2018weakly}. 

Following the same trend, we propose a weakly-supervised approach for surgical tool {\it tracking}. First, we train a surgical tool detector on image-level labels.
From a class peak response, we learn the whole region boundaries of the surgical tools in the laparoscopic videos.
Then, we employ a Convolutional LSTM (ConvLSTM) to learn the spatio-temporal coherence across the surgical video frames. 
Without any spatial appearance and motion cue, the ConvLSTM is naturally able to learn the tools' spatio-temporal positions for tracking.
To the best of our knowledge, this is the first study that builds a complete deep learning tracking model for endoscopic surgery using weak supervision and also the first study that evaluates surgical tool tracking performance on MOT metrics.
Finally, we evaluate our approach on the largest public endoscopic video dataset to date, \emph{Cholec80}, which is fully annotated with binary presence information for 7 tools and of which 5 videos have been annotated with bounding box information for testing.

The remaining of this paper presents a review of related literature (sect.~\ref{sec:literature_review}), our proposed methods (sect.~\ref{sec:methodology}) and implementation details (sect.~\ref{sec:experiments}), followed by a comparative discussion of our results (sect.~\ref{sec:results_discussion}) and a conclusion (sect.~\ref{sec:conclusion}).

\section{Related Work}
\label{sec:literature_review}
In the past, while many works have focused on surgical tools detection \cite{tmi:twinanda2017endonet,ipcai:richa2011visual,miccai:sznitman2014fast,miccai:vardazaryan2018weakly,miccai:sznitman2011unified,mai:al2018monitoring,wacv:jin2018tool}, less have explored their localization \cite{miccai:vardazaryan2018weakly,wacv:jin2018tool,mai:rieke2016real} and tracking \cite{ipcai:richa2011visual,miccai:sznitman2011unified,miccai:sznitman2012data} from video data only. This may be because most localization tasks, and tracking by extension, have been traditionally approached with fully supervised methods that require spatially annotated datasets \cite{mai:bouget2017vision}.

\paragraph{\textbf{Surgical Tools Detection, Localization and Tracking:}}
We review some of the endoscopic tool detection, localization and tracking approaches from the literature, which are mostly concentrated in retinal microsurgery \cite{ipcai:richa2011visual,miccai:sznitman2014fast,miccai:sznitman2011unified,mai:rieke2016real,miccai:sznitman2012data}, and laparoscopic surgery \cite{miccai:sznitman2014fast,miccai:vardazaryan2018weakly,wacv:jin2018tool}. 
In most cases, the localization and/or tracking models rely on a fully supervised object detector \cite{miccai:sznitman2012data,lstm_da:milan2017online,mai:rieke2016real,miccai:sznitman2014fast}. Sometimes, a unified detector-tracker framework is used \cite{miccai:sznitman2011unified}. 
Whereas some tracking models use an optical flow tracker \cite{miccai:lo2003episode}, others have casted tracking as an energy minimization function using a gradient-based tracker \cite{ipcai:richa2011visual,mai:rieke2016real,miccai:sznitman2012data}, density estimation \cite{miccai:sznitman2011unified}, or an image similarity measure based on weighted mutual information \cite{ipcai:richa2011visual}.
From another perspective, works in \cite{miccai:sznitman2014fast,mai:rieke2016real} model the tool articulation parts and estimate the instruments locations by either a non-maximum suppression technique \cite{miccai:sznitman2014fast} or by template tracking \cite{mai:rieke2016real}.
In \cite{wacv:jin2018tool}, a fully supervised region-based convolutional network is employed to detect and localize surgical tools in laparoscopic videos. While the model is able to detect tool presence and localize beyond the tool tips, it requires bounding box annotations for training.
Also, the approach does not take into account the temporal consistency over time. The experiments are carried out on selected images from surgical videos in the \textit{m2cai-tool-locations} dataset.
In all the above-reviewed literature, the object detection and localization models, and, by extension, the trackers, are fully supervised on a spatially annotated dataset for position estimation.

\paragraph{\textbf{Weak Supervision:}}
Considering the difficulty to annotate datasets spatially, \cite{miccai:vardazaryan2018weakly} localized surgical tools on a whole laparoscopic video sequence using a weakly-supervised Fully Convolutional Networks (FCN) model. 
The localization is limited to the center pixels of the tools. 
Other interesting applications of weak supervision in medical imaging are seen in the segmentation of cancerous regions in histopathological images \cite{ieee:jia2017constrained} and in the detection of the region of interest (ROI) in chest X-rays and mammograms \cite{miccai:hwang2016self}.
The aforementioned weakly-supervised approaches do not exploit the temporal coherence of a video sequence and do not perform tracking. 

\paragraph{\textbf{Temporal Coherence:}}
An effort to utilize the temporal interconnection of video frames in deep learning approaches is presented in \cite{cvpr:luo2018fast}, where 3D object detection and motion forecasting are integrated to track moving objects. 
The core idea of this approach is the modeling of temporal coherence using early and late fusion in CNNs. 
However, the decisions on the birth/death of an object track are hard-coded by the aggregation of past, current and future predictions.
A unified approach for processing temporal streams of images is also presented in \cite{arXiv:liu2017mobile}, where ConvLSTM are injected in between convolution layers to refine feature map and propagate frame-level information across time. 
While these models exploit temporal-coherence of a video sequence, the approaches are all fully supervised.
Temporal coherence has also been used to improve binary tool presence detection by adding an LSTM to the output of ResNet-50 \cite{cvpr:mishra2017learning} and to the output of ensembled CNN architectures \cite{mai:al2018monitoring}. These approaches are however not constructed for localization and tracking.

\medskip
\noindent Constructing upon \cite{miccai:vardazaryan2018weakly}, we implement a weakly-supervised approach to train a ConvLSTM for surgical tool tracking. The model is trained on image-level binary labels only. 
Like in \cite{wacv:jin2018tool}, our model localizes the whole region boundaries of the tools beyond their center points.
We leverage the ConvLSTM's spatio-temporal ability to learn the surgical tool trajectories across the frames without requiring more than the image-level class labels.
The ConvLSTM does not only improve presence detection, but also refines and propagates localization feature map across time and helps to track over occlusions. 
Its internal gating mechanism enables it to naturally handle the birth, propagation, and death of tool tracks.

\section{Methodology}
\label{sec:methodology}
\subsection{Architecture}\label{sec:architecture}
Our models are built on the ResNet-18 architecture \cite{cvpr:he2016deep}, which is popular for its excellent performance on object detection. 
We present below the architectures used in this work.

\subsubsection{FCN Baseline}\label{sec:baseline_fcn_model}
\begin{figure}[t]
    \centering
        \includegraphics[width=0.75\textwidth]{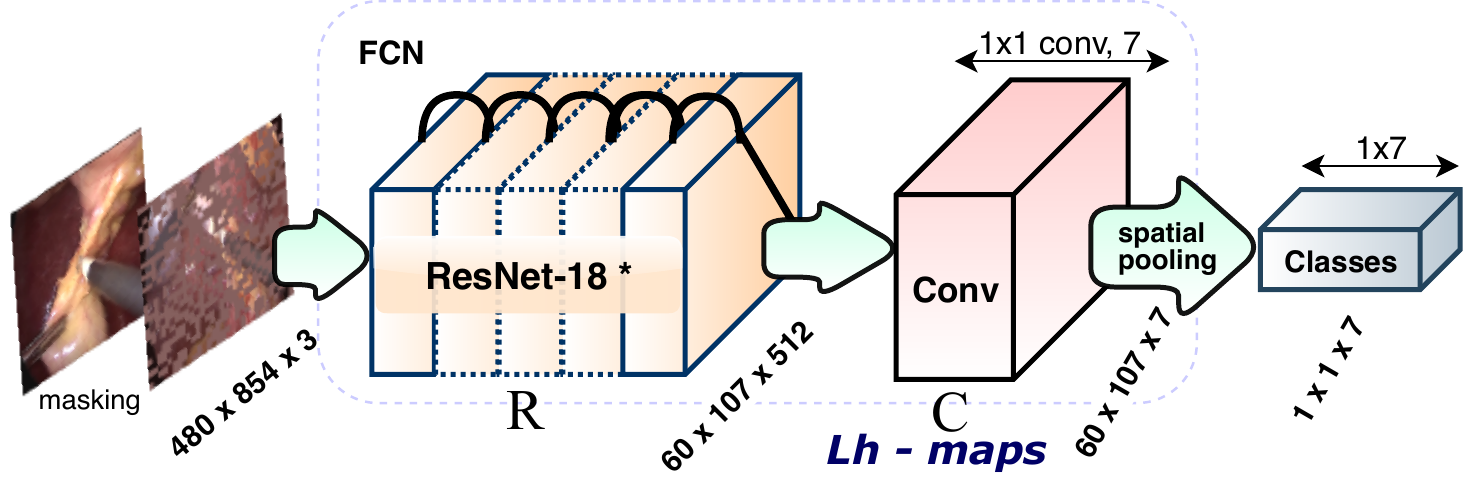}
        \caption{Architecture of FCN baseline model ($\mathbb{R+C}_{M1\_mask}$ variant).}
    \label{fig:fcn_baseline}
\end{figure}

\bd{Detector: }
To build a tracker for surgical tools, we first reproduce the FCN model in \cite{miccai:vardazaryan2018weakly} (illustrated in Fig.~\ref{fig:fcn_baseline}) with similar accuracy on surgical tool presence detection and spatial localization. 
The general configuration of the FCN baseline model is $\mathbb{R+C}$, where $\mathbb{R}$ represents a modified \textit{ResNet-18} network and $\mathbb{C}$ a convolution layer. 
$\mathbb{C}$ is a 1x1 7-channel convolution layer that acts as the \textit{localization heat map (Lh-map)}. 
It replaces the FC-layer of $\mathbb{R}$. The strides of the last two blocks of \textit{ResNet-18} are adjusted from 2 to 1 pixel to obtain an \emph{Lh-map} with higher resolution.
The FCN model is trained only on tool presence binary labels.
Taking an RGB input image, the $\mathbb{R}$ layer extracts spatial feature maps and the $\mathbb{C}$ layer uses $7$ convolution filters to convolve these maps into a $7$-channel \emph{Lh-map}.
Each channel is by design constrained to learn and localize a distinct tool type out of the 7 tools present in the considered laparoscopic procedure.
With wildcat spatial pooling \cite{cvpr:durand2017wildcat}, we transform the \emph{Lh-map} into a $1\times7$ vector of class-wise confidence values indicating the probability of a tool being present or absent. The positive classes are selected by a threshold of 0.5.

Apart from the single-channel map (\textit{single-map}) model (\textit{$\mathbb{R+C}_{M1}$}) discussed above, a multiple channel map (\textit{multi-map} \cite{cvpr:durand2017wildcat}) variant (\textit{$\mathbb{R+C}_{M4}$}) is built by using a convolution layer with \textit{$m\times7$} channels followed by an average pooling over each consecutive group of $m$ channels to give the final 7 channels. 
We retain \textit{$m=4$} as used in \cite{miccai:vardazaryan2018weakly}. 
Both variants are trained on the normal images and on patch masked images (\textit{$\mathbb{R+C}_{M1\_mask}, \mathbb{R+C}_{M4\_mask}$}). 
During patch masking \cite{iccv:singh2017hide}, random patches are created on the original images and their pixel values replaced with the mean pixel value of the entire training dataset. 
According to \cite{iccv:singh2017hide}, this enables the network to learn meticulously the necessary details of the object of interest. 
We now have a total of four variants of the FCN models by pairing the two models based on single- or multi-map with the two models based on masked- or unmasked-input.

\paragraph{\bd{Tracker: }}
We leverage the separation of the tool type in the $7$-channel \emph{Lh-map} from the FCN detector to build a baseline model for tool tracking.
For localization, the raw \emph{Lh-map} is resized to the original input image size by bilinear interpolation. 
Then, with a disc structuring element of size 12, we perform a morphological closing on the resized map to fill small holes in the image.
On each channel of the \emph{Lh-map}, a segmentation mask is extracted from the connected component around the pixel with maximum value using Otsu automatic thresholding \cite{smc:otsu1979threshold}. A bounding box is then drawn over the mask to extract the tool location coordinates.

For tracking, the Intersection over Union (IoU) of the bounding boxes between the current frame $F_t$ and the previous frame $F_{t-1}$ is computed for each detected tool. Tools detected at time $t$ are included in the previous trajectories if the IoU with previous detections at time $t-1$ is at least $0.5$.
In the case of multiple instances of the same tool, the closest tool instance compared to the detections in $F_{t-1}$ is selected.
Unmatched tools are discarded as false detections, while untracked tools are discarded as dead tracks.

\subsubsection{ConvLSTM Tracker}\label{sec:convlstm_tracker}
The aforementioned FCN baseline tracker is trained on images and does not utilize the temporal cues of video data.
This may be a problem when a tool's motion becomes irregular beyond what an IoU of $0.5$ with the previous frame can capture, since the tracking algorithm is hard-coded. 
Knowing that object motion is encoded in temporal information \cite{cvpr:luo2018fast,arXiv:liu2017mobile}, we propose to integrate a temporal model in the previous FCN framework, in a manner that still allows for weakly-supervised training. This results in an elegant end-to-end tracking method that can model the spatio-temporal motion of the tools and also adapt to the various types of motion appearing in a video. 

As temporal model, we propose to use a recurrent neural network (RNN), with the aim to determine the current position of each tool from the input feature map along with information from prior images captured in RNN's state. In designing this architecture, it is necessary to ensure that the overall network can still retain spatio-temporal information for each tool when being trained in a weakly-supervised manner on binary presence data, namely that the localization information per tool is not lost but remains the key information used for predicting the binary presence.

\begin{figure}[t]
    \centering
        \includegraphics[width=0.75\textwidth]{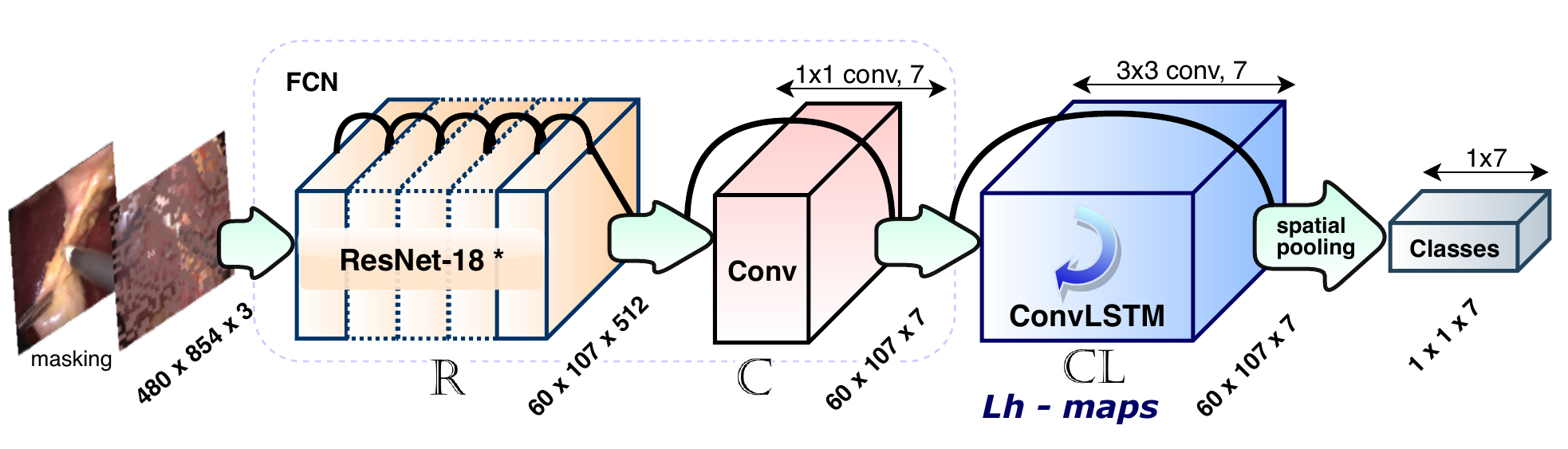}
        \caption{The ConvLSTM tracker architecture with the $\mathbb{R+C+CL}$ configuration.}
    \label{fig:RCCL}
\end{figure}

Using a fully convolutional architecture is key in this regard. We therefore employ a ConvLSTM unit for its ability to learn the \emph{spatio-temporal} dependencies of the localization heat maps. The ConvLSTM achieve this by using a convolution kernel whose receptive field considers temporal information. Compared to stacking a regular LSTM, the spatial relationships are maintained.  And unlike using a simple convolution layer, the ConvLSTM takes into account the features from the previous frames, thereby enforcing consistency across time. At the level of the ConvLSTM, the localization heat maps from each tool remain independent: in this final part of the network, information is indeed not shared across maps to retain the spatial information for each tool. Our \emph{ConvLSTM Tracker} is constructed by adding a ConvLSTM unit to the FCN baseline detector, as illustrated in Figure~\ref{fig:RCCL}. We have explored several variants of the architecture, described further below. 
By naturally smoothing out the class peak activations using temporal information, the ConvLSTM replaces the IoU-based selection from the baseline tracker and naturally handles the birth and death of tracks for each tool. 

In practice, we construct the ConvLSTM trackers using the baseline model $\mathbb{R+C}_{M1\_mask}$, which has the best performance across the 3 tasks (as shown in Tables~\ref{tab:detection}-\ref{tab:tracking}). We select the single-map architecture, as the multi-map architecture is more complex and shows no better performance both in \cite{miccai:vardazaryan2018weakly} and in our baseline spatial experiments (as shown in Tables~\ref{tab:localization} \&~\ref{tab:tracking}). 
Like in ResNet ($\mathbb{R}$), which contains skip connections between its layers, we include skip connections in the $\mathbb{C}$ and $\mathbb{CL}$ layers for their efficiency in training large networks \cite{cvpr:he2016deep,iwwwcsw:wei2018residual}.

To perform weakly supervised training on image-level labels $y$, we transform the \emph{Lh-maps} (see Figure~\ref{fig:RCCL}) into class-wise probabilities $\hat{y}$ using wildcat pooling \cite{cvpr:durand2017wildcat}.
We then learn a weighted cross-entropy loss function $\mathcal{L}$ for multi-label classification:
\begin{equation}\label{eq:loss}
\mathcal{L} \longleftarrow \sum_{c=1}^C\frac{-1}{N}\left [ \mathcal{W}_c y_c \log(\sigma(\hat{y}_c)) + (1-y_c) \log(1-\sigma(\hat{y}_c)) \right],
\end{equation}
where $y_c$ and $\hat{y}_c$ are respectively the ground truth and predicted tool presence for class $c$, $\sigma$ is the sigmoid function, and $\mathcal{W}_c$ the weight for class $c$. 
The effect of the class weights $\mathcal{W}_c$ in this loss function is that $\mathcal{W}_c > 1$ decreases false negatives (FN) while $\mathcal{W}_c < 1$ decreases false positives (FP). With this, we counteract the polarizing effect of class imbalance by reducing FN for less frequent tools and reducing FP for dominant tools. The $\mathcal{W}_c$ is calculated as in Equation~\ref{eq:pos_weight}, where $m$ is the median frequency of all tools in the train set and $F_c$ is the frequency of the tools in class $c$:
\begin{equation}\label{eq:pos_weight}
    \mathcal{W}_c \longleftarrow \frac{m}{F_c}.
\end{equation}

We propose three different configurations with similar architectures:
\begin{enumerate}
\item $\mathbb{R+C+CL}$
\item $\mathbb{R+CL+C}$
\item $\mathbb{R+CL}$
\end{enumerate}
where $\mathbb{R, C}$ and $\mathbb{CL}$ are ResNet, Convolution and ConvLSTM respectively.
\paragraph{$\mathbb{R+C+CL}$ \bd{Configuration}:} In this configuration, illustrated in Figure~\ref{fig:RCCL}, the ConvLSTM receives spatial input features from the $\mathbb{C}$ layer, refines them with temporal information and outputs \emph{spatio-temporal Lh-maps}. The motivation for adding the ConvLSTM unit immediately after the baseline FCN ($\mathbb{R+C}$) is to refine the spatial \emph{Lh-maps} with spatio-temporal information. This helps to smooth the class peak activations as well as the shape and size of the tools segmentation masks. It is important to note that the localization process is performed on the \textit{spatio-temporal Lh-maps}.

\paragraph{$\mathbb{R+CL+C}$ \bd{Configuration}:} With the ConvLSTM unit added before the last Convolution layer of the baseline FCN, it refines the $\mathbb{R}$ spatial features with spatio-temporal information before localization by $\mathbb{C}$. This guides the model in choosing relevant features based on temporal information across the video frames. By doing so, the receptive fields of  $\mathbb{C}$ become aware of the temporal information. 
It is also important to note that the localization is on the $\mathbb{C}$ layer, which receives a spatio-temporal feature map and outputs a \textit{spatial Lh-map}. 
This model is expected to be more robust to occlusion and noise. 

\paragraph{$\mathbb{R+CL}$ \bd{Configuration}:} The last variant replaces the $\mathbb{C}$ layer of the FCN baseline detector with a ConvLSTM ($\mathbb{CL}$) layer. 
Owing to its internal convolution process, the $\mathbb{CL}$ layer takes over the task of localization from the $\mathbb{C}$ layer as well as the refinement of the feature map with temporal information. This results in a less complex architecture with the localization process on the $\mathbb{CL}$ layer that produces \textit{spatio-temporal Lh-maps}.

\section{Experimental Setup}
\label{sec:experiments}
\subsection{Dataset}\label{sec:dataset_analysis}
The dataset used in this experiment is \textit{Cholec80} \cite{tmi:twinanda2017endonet}. It consists of 80 videos of cholecystectomy surgeries aimed at removing the gallbladder laparoscopically, monitored through an endoscope.
The videos are recorded at the frame rate of $25fps$ and downsampled to $1fps$ at which the tool presence binary annotations are generated. While most of the videos are recorded at a resolution of $854\times480$ pixels, a few are $1920\times1080$ pixels with the same aspect ratio. For uniformity, all frames are resized to $854\times480$ pixels in our experiment. 
For the tool detection task, the dataset is split into 40, 10 and 30 videos for training, validation, and testing respectively. 
For localization and tracking evaluation, we use 5 videos from the test set annotated with tool centers and bounding boxes around the tool tips. The tool shafts are excluded, following common practice.

\subsection{Training}\label{sec:training}
All the models presented in this paper are trained by transfer learning. 
The FCN baseline models are trained for 160 epochs with stepwise decaying learning rates starting at the initial values of $1e^{-1}$ and $1e^{-3}$ for the $\mathbb{R}$ and the $\mathbb{C}$ layers respectively. We use the different learning rates to strike a learning balance for $\mathbb{R}$ that has been pretrained on ImageNet and $\mathbb{C}$ that is trained from scratch.

The ConvLSTM and the baseline models have the same backbone feature extractor which converges after 160 epochs. 
For spatial-temporal refinement, $\mathbb{C}$ and $\mathbb{CL}$ of the ConvLSTM models are trained up to 120 epochs with an initial learning rate of $1e^{-3}$ that decays exponentially. 
During this period, the $\mathbb{R}$ layer is frozen for fair comparison with the baseline. 
The training input images are masked by $16\times16$ patches selected randomly at a probability of $0.5$. This patch masking, together with rotation and horizontal flipping of the images, are the 3 data augmentation styles employed in training the $\mathbb{R+CL}$ model. In finetuning the ConvLSTM layer, the dataset augmentation is limited to image patch masking to reduce the training time, since the video dataset already contains lots of variability in the images.

All the models are trained for multi-label classification.
The optimized loss function $\mathcal{L}$ is the weighted cross-entropy with logits presented in Equation~\ref{eq:loss}.
An $L_2$ norm with a weight decay constant of $1e^{-4}$ for the baseline FCN and $1e^{-5}$ for the ConvLSTM models is applied to regularize the optimization.
The models are trained with the momentum optimizer (initial momentum $\mu=0.9$) and using truncated back-propagation.
Owing to our GPU memory constraints and large input dimension, the network is trained with a maximum batch size of 16 and the ConvLSTM models are unrolled for 16 timesteps.
We also propagate the ConvLSTM states \emph{between} batches. To maintain continuity in a video, we initialize the ConvLSTM input states of every batch with the output states of the immediate previous batch.
States propagation is performed during testing as well.
Our model network is implemented in Tensorflow using TFRecords to build the dataset input pipeline and trained on GeForce GTX 1080 Ti GPUs.

\section{Results and Discussion}
\label{sec:results_discussion}
\subsection{Presence Detection Results}
\label{sec:presence_detection}
To quantify the tool presence detection results, we use average precision (AP), which is defined as the area under the precision-recall curve. 
Comparing the AP of our model with the baseline (as presented in Table~\ref{tab:detection}) shows that temporal information is helpful in improving the tool presence detection by over $5.0\%$. The performance improvement can also be seen across the tools. This suggests that the temporal information helps the detection of tools under occlusion and noise. 
\begin{table}[!htbp]
\caption{Tool presence detection average precision (AP) for the evaluated models.}
\setlength{\tabcolsep}{0.66em}
\begin{tabular}{l|l|ccccccc|r}
\vdir{\makecell{}} & Model & \vdir{Grasper} & \vdir{Bipolar} & \vdir{Hook}  & \vdir{Scissor} & \vdir{Clipper} & \vdir{Irrigator} & \vdir{\makecell{Specimen\\Bag}} & \textbf{mAP}  \\ \hline
\rule{0pt}{3ex}  \multirow{4}{*}{\makecell[l]{\vdir{Baseline}}}  
& $\mathbb{R+C}_{M1}$           & 96.7      & 91.9     & 99.4     & 50.6     & 80.3        & 85.2      & 88.3      & 84.6\\
& $\mathbb{R+C}_{M1\_mask}$      &\bd{99.8}  & 92.6     & 99.8     & 85.1     & 96.9        & 60.9      & 78.6      & 87.7\\
& $\mathbb{R+C}_{M4}$           & 95.9      & 89.4     & 99.5     & 69.3     & 85.4        &\bd{89.5}  & 87.1      & 87.9\\
& $\mathbb{R+C}_{M4\_mask}$     & 99.6      & 90.9     & 99.8     & 48.5     & 88.5        & 66.2      & 91.0      & 83.6\\\hline
\rule{0pt}{3ex} \multirow{3}{*}{\makecell[l]{\vdir{Ours}}}   
& $\mathbb{R+C+CL}$        & 99.7      &\bd{95.6}  &  99.8     &  86.9     &\bd{97.5}  &  74.7     &\bd{96.1}  &\bd{92.9} \\
& $\mathbb{R+CL+C}$        &\bd{99.8}  &\bd{95.6}  &\bd{99.9}  &  76.1     &  97.1     &  77.4     &  93.9     &  91.4 \\
& $\mathbb{R+CL     }$        & 99.5      &  93.8     &\bd{99.9}  &\bd{90.3}  &\bd{97.5}  &  65.1     &  74.0     &  88.5 \\
\end{tabular}
\smallskip
\label{tab:detection}
\end{table}

\subsection{Spatial Localization Results}
\label{sec:spatial_localization}
To quantify the network's ability to localize the distinct tools in various frames, we compute the bounding box IoUs between the detected tools and the groundtruths.
This performance measure does not take into account the temporal consistency of the tools across the frames.
However, a localization is only considered to be correct if and only if the $IoU\geq0.5$.
Note that this is stricter than the center-in-bounding box localization metric in \cite{miccai:vardazaryan2018weakly}, which does not takes the IoU into consideration.
The localization results compared with our baseline model is presented in Table~\ref{tab:localization}. 

\begin{table}[!htbp]
\caption{Localization accuracy of tools detected at IoU $\geq0.5$ for the evaluated models.}
\setlength{\tabcolsep}{0.66em}
\begin{tabular}{l|l|ccccccc|r}
\vdir{\makecell{}} & Model & \vdir{Grasper} & \vdir{Bipolar} & \vdir{Hook}  & \vdir{Scissor} & \vdir{Clipper} & \vdir{Irrigator} & \vdir{\makecell{Specimen\\Bag}} & \vdir{Mean}\\ \hline
\rule{0pt}{3ex} \multirow{4}{*}{\makecell[l]{\vdir{Baseline}}} 
& $\mathbb{R+C}_{M1}$      &05.9        &20.5        &34.7        &03.5        &06.4        &\bd{55.1}    &44.4        &24.3 \\
& $\mathbb{R+C}_{M1\_mask}$ &15.5        &10.1        &27.8        &20.0        &13.3        &53.7        &06.4        &21.0 \\
& $\mathbb{R+C}_{M4}$      &05.0        &11.5        &15.5        &25.1        &8.7        &42.5        &14.8        &17.6 \\
& $\mathbb{R+C}_{M4\_mask}$ &08.7        &0.01        &25.6        &20.0        &\bd{20.0}    &49.0        &02.2        &17.9 \\ \hline
\rule{0pt}{3ex} \multirow{3}{*}{\makecell[l]{\vdir{Ours}}}
& $\mathbb{R+C+CL}$     &33.8        &\bd{20.8}        &41.9       &21.1      &12.6        &52.1        &23.8        &29.3 \\
& $\mathbb{R+CL+C}$     &\bd{54.5}    &14.6        &\bd{50.0}      &23.2        &11.8        &53.6        &\bd{60.1}    &\bd{38.2} \\
& $\mathbb{R+CL}$          &42.5        &08.0        &44.4        &\bd{25.3}        &14.0        &53.5        &41.7        &32.8 \\
\end{tabular}
\smallskip
\label{tab:localization}
\end{table}
From this result, our model improved the spatial localization of five out of the seven surgical tools: grasper, bipolar, hook, scissors and specimen bag.
For the irrigator and the clipper, for which the ConvLSTM models do not have the best performance, the performance is comparable.  
Generally, the ConvLSTM shows a good performance on this metric by improving the mean accuracy by $13.9\%$, illustrating the benefits of using temporal information during training. 
Also, all the ConvLSTM models outperform all the baseline models on mean spatial localization accuracy.
This shows that the temporal data modeling can help in understanding the full spatial boundaries of moving objects.

\subsection{Motion Tracking Results}
For the tracking performance evaluation, we adopted the widely used CLEAR MOT metrics \cite{mot:bernardin2008evaluating}: multiple objects tracking precision (MOTP) and multiple objects tracking accuracy (MOTA). 
MOTP, a measure of the localization precision, gives the average overlap between all the correctly matched hypotheses and their corresponding targets for a given IoU threshold ($\Theta$).
\begin{equation}
    MOTP = \frac{\sum_{t,i}D_{t,i}}{\sum_{t}C_t},
    \label{eq:motp}
\end{equation}
where $D_{t,i}$ is the bounding box IoU of the tracked target $i$ with the groundtruth, $C_t$ is the number of matches in frame $t$. The value typically ranges between [$\Theta$\%, 100].
On the other hand, MOTA shows the tracker's ability at keeping consistent trajectories. It evaluates the effectiveness of the tracker from three errors, namely FP, FN and identity switches (IDSW) in respect the number of groundtruth objects (GT) as in equation~\ref{eq:mota}:
\begin{equation}
    MOTA = 1-\frac{\sum_{t}FP_t+FN_t+IDSW_t}{\sum_{t}GT_t}.
    \label{eq:mota}
\end{equation}
The score, which usually ranges between (-$\infty$, 100], can be negative in cases where the number of errors made by the tracker exceeds the number of all objects in the scene. Refer to \cite{mot:bernardin2008evaluating} for more details on the MOT metrics.
\begin{table}[!htbp]
\caption{Tracking performance of the evaluated models.}
\setlength{\tabcolsep}{0.27em}
\begin{tabular}{c|l|cc|cc|cc||cc}
\rule{0pt}{3ex} \multirow{2}{*}{\makecell[l]{\vdir{}}}&
&\multicolumn{2}{c|}{$\Theta=0.3$} &\multicolumn{2}{c|}{$\Theta=0.5$} &\multicolumn{2}{c||}{$\Theta=0.7$}&\multicolumn{2}{c}{Mean}\\
& Model & {MOTP} & {MOTA} & {MOTP} & {MOTA} & {MOTP} & {MOTA} & {MOTP} & {MOTA} \\ \hline
\rule{0pt}{1ex} \multirow{4}{*}{\makecell[l]{\vdir{Baseline}}}
& $\mathbb{R+C}_{M1}$              &58.1        &29.8        &\bd{66.6}    &19.3        &77.3      &05.3        &67.3        &18.1 \\
& $\mathbb{R+C}_{M1\_mask}$        &49.9        &47.9        &61.2        &21.2        &75.3        &02.7        &62.1        &23.9 \\
& $\mathbb{R+C}_{M4}$              &46.6        &29.6        &60.4        &09.6        &75.4        &-00.3        &60.8        &13.1 \\
& $\mathbb{R+C}_{M4\_mask}$        &48.3        &40.4        &61.0        &15.3        &75.8        &01.9        &61.7        &19.2 \\ \hline
\rule{0pt}{1ex} \multirow{4}{*}{\makecell[l]{\vdir{Ours}}}
& $\mathbb{R+C+CL}$            &58.0    	    &46.4        &65.9        &29.4    &\bd{77.4}        &03.2        &67.1       &26.3 \\
& $\mathbb{R+CL+C}$            &\bd{59.0}       &\bd{59.6}    &65.9        &\bd{41.0}    &77.3       &\bd{09.0}    &\bd{67.4}  &\bd{36.5} \\
& $\mathbb{R+CL}$             &54.4             &47.7        &63.3        &26.1        &76.7        &00.3        &64.8        &24.7 \\
\end{tabular}
\smallskip
\label{tab:tracking}
\end{table}

The tracking results across varying $\Theta$ in comparison with our baseline models are presented in Table~\ref{tab:tracking}.
Our approach improved the baseline performance significantly. The results show that with comparable MOTP, ConvLSTM can improve the MOTA baseline by $11.7\%$ at $\Theta=0.3$, $19.8\%$ at $\Theta=0.5$ and $3.7\%$ at a strict $\Theta=0.7$. Generally, the ConvLSTM shows its ability to learn a smoother trajectory by outperforming all the baseline in both mean MOTP and mean MOTA significantly.

\subsection{Qualitative Results}\label{sec:qualitative_results}
\begin{figure}[t]
    \centering
        \includegraphics[width=0.99\textwidth]{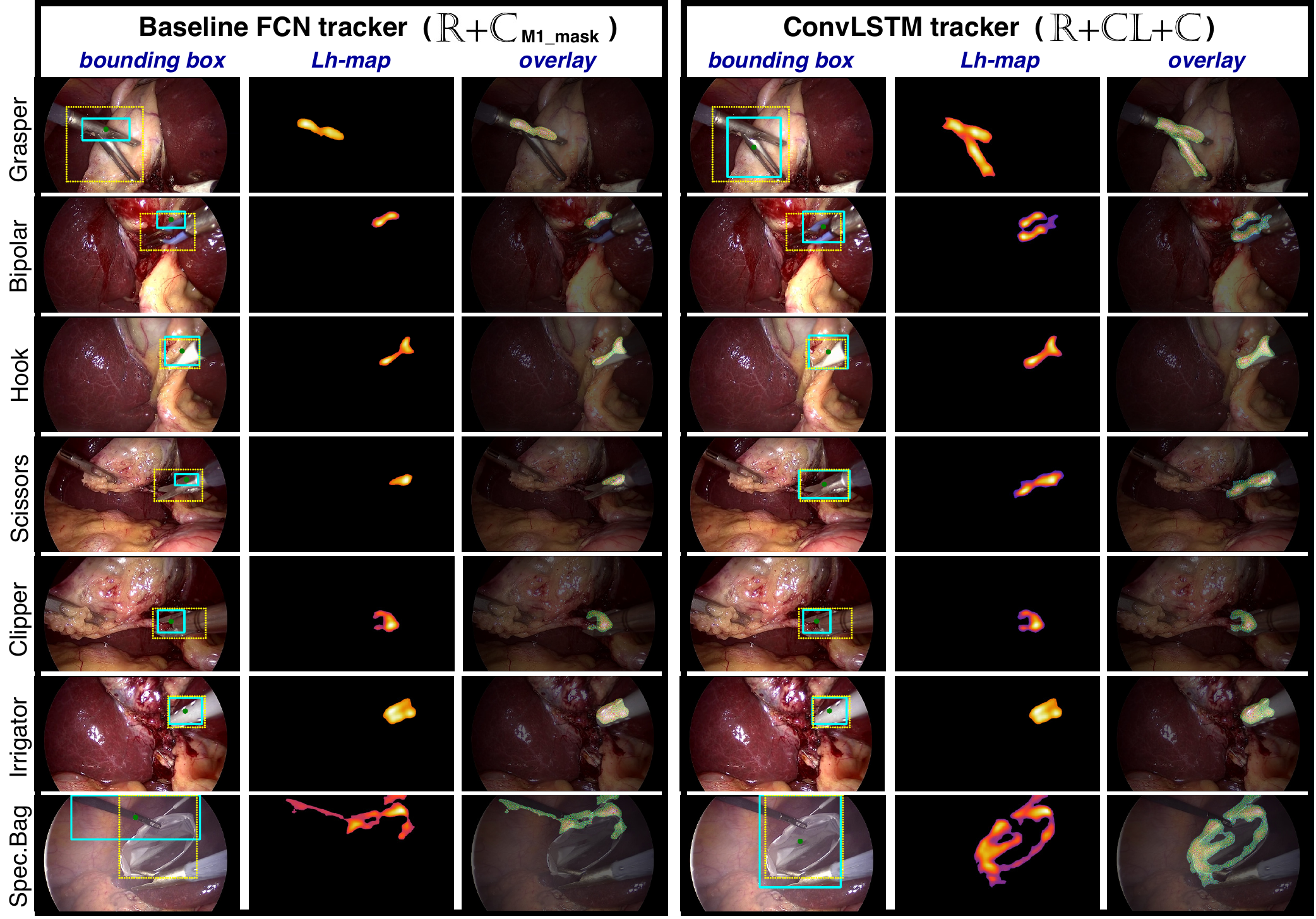}
        \caption{Qualitative results showing the localization and tracking performance of the baseline and ConvLSTM models for the 7 tools. For each tool, we present a comparison of the detected bounding box (cyan in colour) with the ground truth (dotted yellow box), the \emph{Lh-map}, and the overlay of the segmented mask with the original image (best seen in colour).}
        \label{fig:qualitative_results} 
\end{figure}

The qualitative results in Figure~\ref{fig:qualitative_results} show visually how the ConvLSTM is able to leverage the temporal coherence for tracking and localization for the 7 tools. 
From the positioning of the bounding boxes around the tools, it can be seen that the ConvLSTM model learns the region boundaries better than the baseline.
The \emph{Lh-maps} show that the ConvLSTM helps to smooth the localization and approximates the shape and size of the tools in each image. The overlay shows that it satisfactorily learns a trajectory close to the ground truth.
A supplementary video that further demonstrates the qualitative performance of our approach can be found here: \url{https://youtu.be/vnMwlS5tvHE}. 
Our experiments also show that the ConvLSTM model trained on videos at 1fps can generalize to unlabelled videos at 25fps, making it unconstrained by the fps, as can be seen here: \url{https://youtu.be/SNhd1yzOe50}.

\subsection{Discussion}\label{sec:discussion}
The evaluation presented in this paper shows the positive contribution of the ConvLSTMs in modeling temporal data during weakly-supervised training for surgical tool tracking in laparoscopic videos. 
The most notable improvement is seen in the $\mathbb{R+CL+C}$ variant, which has the best results both in localization and in tracking.
We believe that this is due to the fact that in this configuration, $\mathbb{CL}$ refines the feature map from $\mathbb{R}$ with temporal considerations before they are localized separately by $\mathbb{C}$. This is more robust than in $\mathbb{R+C+CL}$ and $\mathbb{R+CL}$, where the temporal refinement at the end of the pipeline may dilute the localization information and output a map with a slightly different semantic. In the $\mathbb{R+CL+C}$ variant, the temporal information across the video frames guides the model in choosing relevant features for the \emph{Lh-maps}. 

In the qualitative results, we observe failure cases in different situations. First, due to the nature of the model, tools are missed when multiple instances of the same class are present. It would be interesting to see if the low activations in the \emph{Lh-maps} could be exploited in order to estimate the number of instances for each class.
The qualitative results also show that the models fail to detect a tool when less than $\frac{1}{5}th$ of its tip is visible. 
We also observe that our models only localize the tool's tip, not its shaft, likely because shafts are similar for all tools and cannot be easily captured by a weakly-supervised approach relying on binary presence.

From the qualitative results, we however notice that the \emph{Lh-maps} produce a weak segmentation of the tool tips, suggesting that this approach could be extended to segmentation.

\section{Conclusion}\label{sec:conclusion}
This paper aims at tracking tools in laparoscopic surgical videos without using any spatial annotation during training. 
A weakly-supervised Convolutional LSTM approach that relies solely on binary tool presence information is proposed. 
First, we build a baseline tracker by performing a one-to-one data association on the localization results generated by the FCN proposed in \cite{miccai:vardazaryan2018weakly}.
Then, we propose a fully convolutional spatio-temporal model for end-to-end tracking that is suitable for weakly-supervised training. It relies on a ConvLSTM that leverages the temporal information present in the video to smooth the class peak activations and better detect the presence of tools, optimize their spatial localization and smooth their trajectory over time. 
This approach is evaluated on the Cholec80 dataset and yields $12.6\%$ overall improvement on MOTA, $13.9\%$ improvement on localization mean accuracy and $5\%$ improvement on tool presence detection mAP.
The quantitative and qualitative results also suggest that the proposed approach could be integrated into a surgical video labeling software to initialize the tool annotations, such as their bounding boxes and segmentation masks.

\begin{acknowledgements}
This work was supported by French state funds managed
within the Investissements d\textsc{\char13}Avenir program by BPI France (project CONDOR) and by the ANR (references ANR-11-LABX-0004 and ANR-10-IAHU-02). 
\end{acknowledgements}

\end{document}